\documentclass[pdflatex,sn-basic]{sn-jnl}

\usepackage{graphicx}%
\usepackage{multirow}%
\usepackage{amsmath,amssymb,amsfonts}%
\usepackage{amsthm}%
\usepackage{mathrsfs}%
\usepackage[title]{appendix}%
\usepackage{xcolor}%
\usepackage{textcomp}%
\usepackage{manyfoot}%
\usepackage{booktabs}%
\usepackage{algorithm}%
\usepackage{algorithmicx}%
\usepackage{algpseudocode}%
\usepackage{listings}%

\graphicspath{{figures/}}
\newcommand{\ex}[1]{\textit{#1}}

\begin{document}

\title[Observatorio L\'azaro]{Observatorio L\'azaro: A self-populating database of anglicism usage in the Spanish press}

\author*[1]{\fnm{Elena} \sur{Álvarez-Mellado}}\email{elena.alvarezm@uam.es}

\affil[1]{\orgdiv{Department of Linguistics}, \orgname{Universidad Autónoma de Madrid}, \orgaddress{ \country{Spain}}}

\abstract{This paper describes Observatorio L\'azaro, a language resource that monitors unassimilated lexical borrowings (predominantly English lexical borrowings or \emph{anglicisms}) in the Spanish digital press. Since April 2020 the system has automatically processed the daily output of a collection of news outlets, detected borrowings with a neural sequence-labeling model, and made the results available through a public web interface and API. The result is a continuously updated diachronic database which, at the time of writing, records more than two million borrowings across 1.88~million articles and 993~million running tokens of text (2020--2026).  The paper documents the resource: we describe the end-to-end pipeline (acquisition, detection, post-processing, storage and access), the data model and the terms of availability; we evaluate the resource through the detector's held-out performance (span-level F1~=~0.86 for the borrowing class), inter-annotator agreement on the training corpus (Cohen's $\kappa$~=~0.91) and a manual precision audit of 1{,}000 spans from the deployed data; and we situate it with respect to Spanish borrowing lexicography, annotated borrowing corpora and neology-monitoring observatories in other languages. The data shows that unassimilated anglicisms are used in the Spanish press at a frequency of approximately two anglicisms per thousand tokens, and that this rate remains stable. Our statistical analysis over six years reveals that the anglicism vocabulary in Spanish behaves as an open and growing class, with 58.7\% of its types attested only once (53.6\% after correcting for detection precision), and that its density is highest in the fashion, technology and lifestyle sections and lowest in political and institutional news. The resource is intended to complement static borrowing dictionaries and one-off annotated corpora by providing a continuously updated record of borrowing in the Spanish press.}


\maketitle

\section{Introduction}\label{sec:intro}

Multilingualism and contact between languages have been the norm throughout history rather than the exception. As \citet{thomason_sarah_g_social_2003} puts it, ``all languages are mixed in a weak sense: there is no natural human language in which foreign material is wholly lacking''. Contact is in fact one of the major inducers of language change: speakers take patterns belonging to one language and incorporate them into another, a process known as linguistic borrowing \citep{haugen_analysis_1950,weinreich_languages_1963}. When what is transferred belongs to the lexicon, we speak of \emph{lexical borrowing}.

Lexical borrowing is a particularly informative object of study because it sits at the intersection of the social and the systemic. A word may be imported together with an artefact or a technique, documenting a cultural exchange between communities; or it may be imported because the foreign form is perceived as more prestigious than the native one, reflecting a social hierarchy that shapes language use. Borrowings are thus evidence of contact between languages and of the social dynamics between the communities that speak them \citep{zenner_cognitive_2012}. At the same time, which words get borrowed, what status they acquire, and how far they are adapted to the recipient language reveal the grammatical patterns and speaker expectations that operate in that language, and the general constraints on language change \citep{van1994modeling,haspelmath_ii_2009}.

The borrowing of English vocabulary into other languages (words such as \ex{influencer}, \ex{streaming}, \ex{podcast} and \ex{look}) is the most salient instance of this process in contemporary European languages \citep{furiassi_anglicization_2012}. Readers of the French press were estimated to encounter a new lexical borrowing roughly every thousand words, with English borrowings outnumbering all other donor languages combined \citep{chesley_predicting_2010,chesley_lexical_2010}. Spanish is no exception to this \citep{nunez_nogueroles_up--date_2017}: in the Chilean press, borrowings were found to account for some 30\% of neologisms, 80\% of them anglicisms \citep{gerding_anglicism_2014}; and for European Spanish, anglicisms were estimated to make up around 2\% of the vocabulary used in the newspaper \ex{El Pa\'is} in 1991 \citep{rodriguez_gonzalez_spanish_2002}, a figure that is likely to be considerably higher today, but that no continuously updated measurement has been able to confirm.

Existing estimates of how pervasive lexical borrowings are are scarce, dated and hard to compare, as studying the phenomenon empirically has traditionally been difficult. The lexicographic tradition documents borrowings authoritatively but statically and with a lag of years \citep{pratt1980,lorenzo1996,rodriguezlillo1997,rodriguez2017}. Corpus-based studies capture a synchronic slice and, crucially, have depended on the manual inspection of general reference corpora, which forces researchers either to annotate an entire corpus by hand or to restrict the enquiry to a predefined list of query terms (Section~\ref{sec:related-corpus}). Neither strategy scales to an on-going process: borrowings are notoriously volatile \citep{poplack2017borrowing}, and because they are also a sparse phenomenon, exhaustive repertoires require very large corpora, which is not feasible without automation. 

There has been a recent wave of computational work on borrowing detection, which has produced annotated datasets and models \citep{alvarezmellado2020headlines,adobo2021,alvarezmellado2022,ahmadi2025conloan}, but not a standing observation of the phenomenon as it unfolds. What has been lacking is a resource that combines four properties: large scale, diachronic depth, open availability and continuous updating, that is, a standing record of which English forms enter the Spanish press, when, where and how often.

In this paper we describe Observatorio L\'azaro\footnote{\url{https://observatoriolazaro.es}} \citep{alvarezmellado2020obs}, an automatic monitoring system that, every day since April 2020, retrieves the articles published by a collection of Spanish news outlets, detects the unassimilated borrowings they contain with a neural sequence-labeling model, and publishes the resulting data through a public website and API. Its by-product is a cumulative diachronic database that, as of July 2026, records 2{,}007{,}647 borrowing tokens in 1{,}880{,}377 articles spanning 993.4~million running tokens. 

The aim of the resource is to make the analysis of anglicism usage possible in real context, at large scale, and in a systematic and data-driven fashion, so as to support both lexicographic work and corpus-linguistic research. The detection model and the annotated corpus on which it was trained were introduced in prior work \citep{alvarezmellado2022}; the contribution of the present paper is the operational monitoring system and the diachronic resource built on top of that model: its pipeline, its data model, its terms of availability and reuse, its evaluation as a resource, and a first characterization of what six years of continuous monitoring reveal, including a measurement of anglicism density that the estimates cited above could not provide.

It may be useful to state plainly what this adds to the work it builds on. \citet{alvarezmellado2022} introduced the \textsc{coalas} corpus and a set of models trained on it, and evaluated those models on held-out annotated text; that paper is about an annotated dataset and a modeling task. The present paper is about neither. Its object is the system that has been running the resulting model in production since 2020 and the database that six years of doing so have produced: the acquisition and preprocessing pipeline, the data model and the access layer (Section~\ref{sec:pipeline}); the conditions under which the data may be reused (Section~\ref{sec:resource-avail}); an evaluation of what the detector does when applied to unfiltered newswire rather than to curated test text, which turns out to differ substantially from the benchmark figure (Section~\ref{sec:eval-audit}); and a characterization of the accumulated output (Section~\ref{sec:overview}). None of these is present in the earlier work, and the deployed evaluation in particular could not have been carried out until the system had run long enough to produce a database worth auditing. A model and an annotated corpus are inputs to a resource of this kind; they are not themselves the resource.

The paper is organized as follows. Section~\ref{sec:related} defines the phenomenon the resource tracks, reviews the descriptive, corpus-based and computational literature on anglicisms in Spanish, and positions L\'azaro among comparable resources. Section~\ref{sec:pipeline} documents the end-to-end pipeline. Section~\ref{sec:resource} describes the resulting database: its data model, size, coverage and its availability and licensing. Section~\ref{sec:eval} evaluates the resource along three axes: model performance, annotation reliability and a manual precision audit of the deployed data. Section~\ref{sec:overview} characterizes the resource statistically over 2020--2026. Section~\ref{sec:apps} discusses applications and reuse; Section~\ref{sec:limits} states limitations.

\section{Background and related resources}\label{sec:related}

\subsection{Lexical borrowing: scope of the phenomenon}\label{sec:related-scope}

Linguistic borrowing is the process of reproducing in one language elements and patterns that come from another \citep{haugen_analysis_1950}, and has been studied extensively within contact linguistics \citep{weinreich_languages_1963}. Several typologies have been proposed to characterize borrowings according to the levels of language involved and the degree of integration of the borrowed element \citep{thomason_language_1992,matras_grammatical_2007,haspelmath_loanwords_2009}. Lexical borrowing in particular involves the incorporation of single lexical units, usually accompanied by morphological and phonological modification to conform to the patterns of the recipient language \citep{poplack_social_1988,onysko_anglicisms_2007}. In Spanish, for instance, borrowed verbs must take the suffix \ex{-ar} or \ex{-ear} to enter the verbal paradigm (\ex{tuitear}, from \ex{to tweet}), and phonological adaptation may surface in spelling, so that the unadapted \ex{spoiler} coexists with the adapted \ex{esp\'oiler}.

All these phonological, orthographic and morphological transformations can lead to a borrowing eventually becoming fully assimilated into the recipient language lexicon, which may lead to native speakers losing the perception of the term being ``foreign'' \citep{lipski_code-switching_2005}. Some authors establish the need of a borrowing being recognized as foreign by native speakers in order to be considered as such \citep{zenner_cognitive_2012}. For instance, a word like \ex{bar} was originally borrowed from English into Spanish, but it has been so assimilated that it is now perceived as a native word by monolingual speakers of Spanish, and its English nature is only seen as etymological. On the other hand, a word like \ex{whisky} (that has also been used in Spanish for some time) has never been fully assimilated and is perceived as a foreign word.

In this paper we deal with unassimilated lexical borrowings in Spanish, that is, words from other languages that have been incorporated into Spanish but that have not undergone the process of assimilation. We focus on lexical borrowings from English (also known as anglicisms), as they represent the vast majority of current borrowings in Spanish.

A further boundary is the one between borrowing and codeswitching, the alternation between two or more languages in discourse that is typical of bilingual communities \citep{poplack_sometimes_1980}. Unlike borrowings, codeswitches are by definition not integrated into the recipient language and do not produce a change in its lexicon; as \citet{haspelmath_ii_2009} puts it, codeswitching ``is not a kind of contact-induced language change, but rather a kind of contact-induced speech behavior''. A clear-cut distinction nonetheless remains elusive, particularly for non-integrated foreign items appearing in otherwise monolingual contexts (so-called ``lone other-language items''). Some authors treat borrowing and codeswitching as a continuum with a fuzzy frontier \citep{clyne_dynamics_2003}, whereas \citet{poplack_myths_2012} argue that they are distinct phenomena and that integration is abrupt rather than gradual \citep{poplack_borrowing_1984,poplack_social_1988}. Criteria proposed to separate them include frequency \citep{stammers_testing_2012}, level of integration \citep{poplack_borrowing_1984}, length \citep{calude_modelling_2020}, speaker bilingual competence \citep{pfaff_constraints_1979} and listedness \citep{treffers-daller_simple_2023}; the debate is unresolved \citep{lipski_code-switching_2005}, and some authors accordingly prefer agnostic terminology such as ``donor-language items'' \citep{deuchar_english-origin_2016} or code-mixing \citep{muysken_bilingual_2000}. As \citet{poplack_myths_2012} put it, ``distinguishing codeswitching and borrowing is the thorniest issue in the field of contact linguistics today''.

For the purpose of this work we follow the approach taken by \citet{poplack_myths_2012} and consider codeswitching and borrowing as two separate phenomena. We consider that codeswitches are fluent multiword interferences that normally comply with grammatical restrictions in both languages and that are produced by bilingual speakers in bilingual discourses (usually in spoken language), while we define lexical borrowings as words of foreign origin that are used by monolingual individuals without knowledge of the donor language.

\subsection{Anglicisms in Spanish}\label{sec:related-es}

The influence of English on Spanish has been a sustained topic of linguistic research for decades \citep{pratt1980,furiassi_anglicization_2012,nunez_nogueroles_up--date_2017}. 
Prior work has proposed different analysis and classifications of anglicism usage in Spanish \citep{lorenzo1996,medina_lopez_anglicismo_1998,rodriguez_gonzalez_anglicisms_1999,nunez_nogueroles_comprehensive_2018} and examined orthographic integration \citep{nunez_nogueroles_typographical_2017}, diachronic shifts \citep{gimeno_menendez_desplazamiento_2003}, typological characteristics \citep{gomez_capuz_towards_1997}, syntactic anglicisms \citep{rodriguez_medina_anglicismos_2002}, sociocultural dimensions \citep{gomez_capuz_prestamos_2004} and lexicographic coverage \citep{balteiro_reassessment_2011}. This tradition is crystallised in reference works such as the \textit{Nuevo diccionario de anglicismos} \citep{rodriguezlillo1997} and the \textit{Gran diccionario de anglicismos} \citep{rodriguez2017}. Authoritative as they are, such works are static: they describe a curated inventory at a point in time, and cannot track frequency, diffusion, or the constant churn of ephemeral borrowings that a living press produces.

\subsection{Limitations of corpus-based study of anglicisms}\label{sec:related-corpus}

Empirical research on anglicism usage in Spanish has been informed chiefly by corpus linguistics \citep{rodriguez_medina_anglicismos_2002,gimeno_menendez_desplazamiento_2003,balteiro_reassessment_2011,nunez_nogueroles_corpus-based_2018}, relying either on general reference corpora such as CREA\footnote{\url{https://corpus.rae.es/creanet.html}} and CORPES\footnote{\url{https://www.rae.es/corpes/}} or on tailor-made corpora built for a specific genre or variety. The main limitation of these corpus-assisted projects is that the retrieval of anglicisms and their concordances relies on the manual lookup of the corpus. That implies that either the whole corpus is processed and annotated by a human (in order to identify all anglicisms it contains), or the research is limited to a list of predefined terms of interest to be queried. This approach seems insufficient to account for an on-going phenomenon like anglicism incorporation, because borrowings are notorious for being volatile \citep{poplack2017borrowing}. In addition, as lexical borrowings are a sparse phenomenon, large corpora are required in order to collect exhaustive repertoires of anglicisms in use, something that is simply not feasible without efficient automatization.

Some corpus interfaces do allow searching for borrowings in general, such as CORPES. Their results, however, leave a lot to be desired. At the time of writing, the concordances obtained by CORPES when querying for borrowings include true borrowings, along with foreign proper nouns (such as \ex{Spice Girls}, \ex{Modern Family} or \ex{Madama Butterfly}), unusual Spanish words (such as \ex{chundachunda} or \ex{fu}, in the idiom \ex{ni fu ni fa}), ill-tokenized words or complete Latin or English phrases (which would correspond more to the phenomenon of codeswitching than borrowing).

Ultimately, both the dissatisfaction with the CORPES anglicism results and the obvious limitations of having to manually process the corpus point in the same direction: the need for robust computer-assisted methods in corpus linguistics in general and in borrowing studies in particular that can facilitate data-driven research on language contact, language change and the lexicon. The resource presented here is a direct response to that need.

\subsection{Automatic borrowing detection}\label{sec:related-detection}

The task of extracting unassimilated anglicisms from Spanish text is a more challenging undertaking than it might appear to be at first. To begin with, lexical borrowings can be single or multiword expressions (e.g., \ex{prime time}, \ex{tie break} or \ex{machine learning}). Second, linguistic assimilation is a diachronic process and, as a result, what constitutes an unassimilated borrowing is not clear-cut. For example, words like \ex{bar} or \ex{club} were unassimilated lexical borrowings in Spanish at some point in the past, but have become so widespread and frequent that the process of phonological and morphological adaptation is now complete and they cannot be considered unassimilated borrowings anymore. In addition, not all English words that appear in Spanish texts will be anglicisms: proper names, literal quotations and other language-mixing phenomena can naturally occur in Spanish text, without any of them being true examples of lexical borrowing. To make things worse, there are non-related words that exist both as English and Spanish words (\ex{quince}, \ex{come}, \ex{primer}, \ex{pie}). These words may or may not be a borrowing depending on the context they appear in. For instance, the word \textit{pie} will be a native word in Spanish when it means ``foot'', but it will be a borrowing in \textit{pie de limón} (``lemon pie''). Similarly, the word \textit{cash} will be a proper noun when referred to singer \textit{Johnny Cash}, but it will be a borrowing in \textit{Estoy sin cash} (``I have no cash'').
All these subtleties make the automatic retrieval of lexical borrowings non-trivial. Consequently, in prior work on anglicism extraction from Spanish text, plain dictionary lookup produced very limited results, with F1 scores of 0.47 \citep{serigos_applying_2017} and 0.26 \citep{alvarezmellado2020headlines}.

Beyond Spanish, computational approaches to borrowing and foreign-word detection have been explored for a range of languages and purposes, from lexicographic and corpus work \citep{hofland_self-expanding_2000,furiassi_retrieval_2007,losnegaard_data-driven_2012,chesley_lexical_2010,serigos_applying_2017} to speech and cross-lingual applications \citep{mansikkaniemi_unsupervised_2012,leidig_automatic_2014}, and more recently with neural and multilingual models \citep{miller_using_2020,nath_generalized_2022,pugh_itml_2023,dinu_robocop_2023}; related work in historical linguistics addresses loanword detection in etymological databases \citep{list_automated_2019,haspelmath_2014_11137}. 

For Spanish specifically, the line of work on which this resource rests reframes the problem as sequence labeling, in which relevant spans (either single-word or multiword) are extracted from sentences, much as in named-entity or multiword-expression recognition works. \citet{alvarezmellado2020headlines} released an annotated corpus of emerging anglicisms in Spanish newspaper headlines; the ADoBo shared task \citep{adobo2021} benchmarked systems for the automatic detection of unassimilated borrowings in the Spanish press; and \citet{alvarezmellado2022} introduced \textsc{coalas}, a manually annotated corpus of 372{,}701 tokens of Spanish newswire containing 3{,}161 unassimilated-borrowing annotations (3{,}038 English and 123 other-language; 1{,}683 unique types), together with a suite of models. The best of these, a BiLSTM-CRF combining bilingual and sub-word embeddings, is the detector deployed in L\'azaro (Section~\ref{sec:pipeline-detect}). These are the immediate predecessors of the present resource: they provide the annotation scheme, the training data and the model, whereas L\'azaro turns them into a deployed, longitudinal resource.

Automatic borrowing detection is also of interest beyond linguistics: borrowings and neologisms contribute to out-of-vocabulary words, which can degrade model performance \citep{zheng_neo-bench_2024}, and their identification has proved relevant for parsing \citep{alex_automatic_2008}, text-to-speech synthesis \citep{leidig_automatic_2014} and as a bootstrapping technique in machine translation for low-resource languages that borrow heavily \citep{tsvetkov_constraint-based_2015,tsvetkov_cross-lingual_2016}.

\subsection{Neology and borrowing observatories}\label{sec:related-obs}

The closest analogues to the resource we introduce are neology and borrowing monitoring systems built on ``monitor corpora'' (corpora that grow over time so that new usage can be observed as it appears). Notable examples include the Observatori de Neologia (OBNEO/BOBNEO) at the Universitat Pompeu Fabra, a long-running project on Catalan and Spanish neology \citep{obneo2004}; large news-monitoring corpora such as the News on the Web (NOW) corpus \citep{davies2013}, which is diachronic and continuously extended but not specialised for borrowing; and Néoveille, a multilingual web platform for tracking and documenting neologisms whose working languages include Spanish \citep{cartier2017}, although the project seems to be discontinued and its site was not reachable at the time of writing. Table~\ref{tab:compare} situates L\'azaro with respect to these resources. It differs from them in combining four properties: it is specialised for unassimilated borrowing detection, fully automatic and updated daily, openly available together with a public API, and now six years deep. We are not aware of another Spanish resource that combines all four, although each is present individually in one or more of the resources listed.

\begin{table}[h]
\centering
\caption{Observatorio L\'azaro among related resources (schematic comparison).}\label{tab:compare}
\begin{tabular}{@{}p{2.9cm}p{1.9cm}p{1.9cm}p{2.0cm}p{2.2cm}@{}}
\toprule
Resource & Target & Method & Diachronic / updated & Access \\
\midrule
Anglicism dictionaries \citep{rodriguezlillo1997,rodriguez2017} & ES anglicisms & manual, lexicographic & static editions & print \\
\textsc{coalas} \citep{alvarezmellado2022} & ES unassim.\ borrowings & manual annotation & static snapshot & open (GitHub) \\
OBNEO / BOBNEO \citep{obneo2004} & CA/ES neologisms & semi-automatic & periodic & partly open \\
N\'eoveille \citep{cartier2017} & multiling.\ neologisms & automatic + expert & continuous & platform (unavailable)\\
NOW \citep{davies2013} & EN news language & corpus, general & continuous & query interface \\
Observatorio L\'azaro & ES unassim.\ borrowings & fully automatic & daily, since 2020 & open web + API + library \\
\botrule
\end{tabular}
\end{table}

\section{The Observatorio L\'azaro pipeline}\label{sec:pipeline}

\subsection{Overview}\label{sec:pipeline-overview}

The main component of Observatorio L\'azaro is an automatic pipeline that extracts anglicisms from Spanish journalistic texts. This pipeline consists of three steps: article retrieval, anglicism extraction and anglicism storage. The output of the pipeline is then aggregated and published through the project website and API (Section~\ref{sec:pipeline-access}).

The observatory monitors the type of borrowings that the model can extract: unassimilated lexical borrowings, with a special focus on unassimilated anglicisms. Consequently, other borrowings such as adapted borrowings (words whose spelling has been modified to comply with the morphophonological and orthographic patterns of the Spanish language, like \ex{f\'utbol} from \ex{football}) or assimilated borrowings (words that have already been registered in reference dictionaries without any italics or quotations, such as \ex{bar}) are not expected to be extracted by the model and are therefore not tracked by the observatory. Similarly, other phenomena like semantic calques, syntactic anglicisms, acronyms and proper names are also considered beyond the scope of the project and are therefore not considered by the observatory. Residual codeswitches and partially assimilated borrowings nevertheless surface among the detection errors (Section~\ref{sec:eval-audit}).

The observatory started monitoring 8 Spanish newspaper sources in 2020. The collection of sources tracked was expanded in September 2022. Table~\ref{tab:throughput} gives the daily throughput of the pipeline in its two configurations. In its initial period the system processed some 640 articles and 195{,}000 tokens a day from 8 sources, extracting around 440 borrowing occurrences (180 distinct) and roughly 19 previously unseen borrowings. After the 2022 expansion it processes some 890 articles and 566{,}000 tokens a day, extracting around 1{,}110 occurrences (430 distinct) and some 35 new borrowings daily.

\begin{table}[h]
\centering
\caption{Average daily throughput of the pipeline, before and after the mid-2022 expansion. The boundary is the outlet expansion of September 2022, and averages are taken over days on which at least one article was retrieved (964 and 1{,}423 days respectively). ``Distinct'' is the mean number of distinct lemmas detected in a day; ``new'' counts lemmas attested for the first time.}\label{tab:throughput}
\begin{tabular}{@{}lrr@{}}
\toprule
Per day (average) & 2020--2022 & 2022--present \\
\midrule
Sources                 & 8 & 26 \\
Articles                & 640 & 890 \\
Tokens                  & 195{,}000 & 566{,}000 \\
Borrowings (all)        & 440 & 1{,}110 \\
Borrowings (distinct)   & 180 & 430 \\
Borrowings (new)        & 19 & 35 \\
\botrule
\end{tabular}
\end{table}

\subsection{Source acquisition}\label{sec:pipeline-acq}

Every day the system retrieves the articles published by a collection of Spanish news outlets (Appendix~\ref{app:outlets}), including the most popular general national dailies (\ex{El Pa\'is}, \ex{El Mundo}, \ex{La Vanguardia}, \ex{elDiario.es}, \ex{ABC}, \ex{20minutos}, \ex{El Confidencial}), a news agency (\ex{Agencia EFE}), the economic press (\ex{El Economista}, \ex{Cinco D\'ias}), sports (\ex{Marca}), and specialist magazines (\ex{Elle}, \ex{Fotogramas}, \ex{Rolling Stone}, \ex{Men's Health}, \ex{Muy Interesante}).

The starting point of the pipeline is an automatic system that connects daily to the RSS feeds of several major news sources from Spain. Every day, the pipeline connects to these sources and extracts the texts of the articles published within the last 24 hours. For each article the system stores the URL, headline, publication date, outlet, section and a running-token count; the article text is passed to the detector. Duplicate articles are avoided by keeping a persistent record of the URLs already scraped, so that each article is processed exactly once. The eight outlets covering general national news have been monitored since  2020 ; a further group was added in September 2022 (Section~\ref{sec:limits}), so composition-controlled longitudinal analyses restrict attention to the stable ``core'' set.

\subsection{Text preprocessing}\label{sec:pipeline-pre}

These articles are then preprocessed (for HTML tag removal, social media embeds, cookie notices, navigation menus and similar elements) and made ready for the next step of the system. The resulting text is then segmented into sentences and tokenized prior to detection using \texttt{spaCy} \citep{ines_montani_2023_10009823}. For every detected borrowing the system stores the surrounding sentence as \emph{context}, the borrowing's start and end token offsets, and two Boolean flags indicating whether the borrowing occurs in a headline and whether it falls within quotation marks or italics.

\subsection{Borrowing detection}\label{sec:pipeline-detect}

Detection is framed as BIO sequence labeling over the tokenized text. The model deployed since August 2022 is the best system of \citet{alvarezmellado2022}: a BiLSTM-CRF that combines bilingual Spanish--English word embeddings with sub-word representations (byte-pair-encoding and character-level embeddings), trained on the manually annotated \textsc{coalas} corpus. This architecture was adopted because it outperformed other alternatives, such as fine-tuning multilingual and Spanish Transformer encoders (mBERT, BETO and XLM-RoBERTa), which did not yield higher results \citep{alvarezmellado2022,delarosa2021}. More recently, large language models have likewise produced only modest results on anglicism and loanword detection, failing to surpass fine-tuned Transformers \citep{sousaahmadi2026,iberbench2025,alvarezmellado2025}. 

The current configuration achieves a span-level F1 of 0.86 for borrowing detection on the \textsc{coalas} test set (Section~\ref{sec:eval-model}). The model also predicts a per-span language label (English vs.\ other); in practice the other-language class is rarely predicted (Section~\ref{sec:eval-model}), so nearly all detections are labeled English and the resource functions as a monitor of anglicisms. From April 2020 to August 2022 an earlier Conditional Random Field model was used \citep{alvarezmellado2020headlines}; the model change is one component of the mid-2022 discontinuity discussed in Section~\ref{sec:limits}. 

The trained model is distributed publicly via \texttt{HuggingFace}\footnote{\url{https://doi.org/10.57967/hf/11427}} and packaged as an installable Python library, \texttt{pylazaro}\footnote{\url{https://pylazaro.readthedocs.io/en/latest/}}, so that third parties can run the same detector over their own text.

\subsection{Post-processing and storage}\label{sec:pipeline-post}

Each detection is (i)~assigned a \emph{lemma} using the \texttt{Pattern}\footnote{\url{https://github.com/clips/pattern}} library \citep{de2012pattern}, run in English mode, since the forms to be normalized are of English or English-like origin and inflect on English rather than Spanish patterns, (ii)~tagged with the language label predicted by the detector (English vs.\ other; reliable only for English, Section~\ref{sec:eval-model}), and (iii)~collected and stored in a MySQL database. For every anglicism, the date, context, newspaper and link to the article where the anglicism was found are stored. In parallel, the system maintains a per-form \emph{index} that aggregates, for each borrowing type, its total number of occurrences, its first-attestation date and a hapax flag.

\subsection{Access layer: website, API and library}\label{sec:pipeline-access}

The output of the pipeline is aggregated and published daily at \url{https://observatoriolazaro.es}. The site is the primary point of access to the resource, and since a substantial part of what the observatory offers is available only through it, we describe it here in some detail.

\begin{figure}
    \centering
    \includegraphics[width=\linewidth]{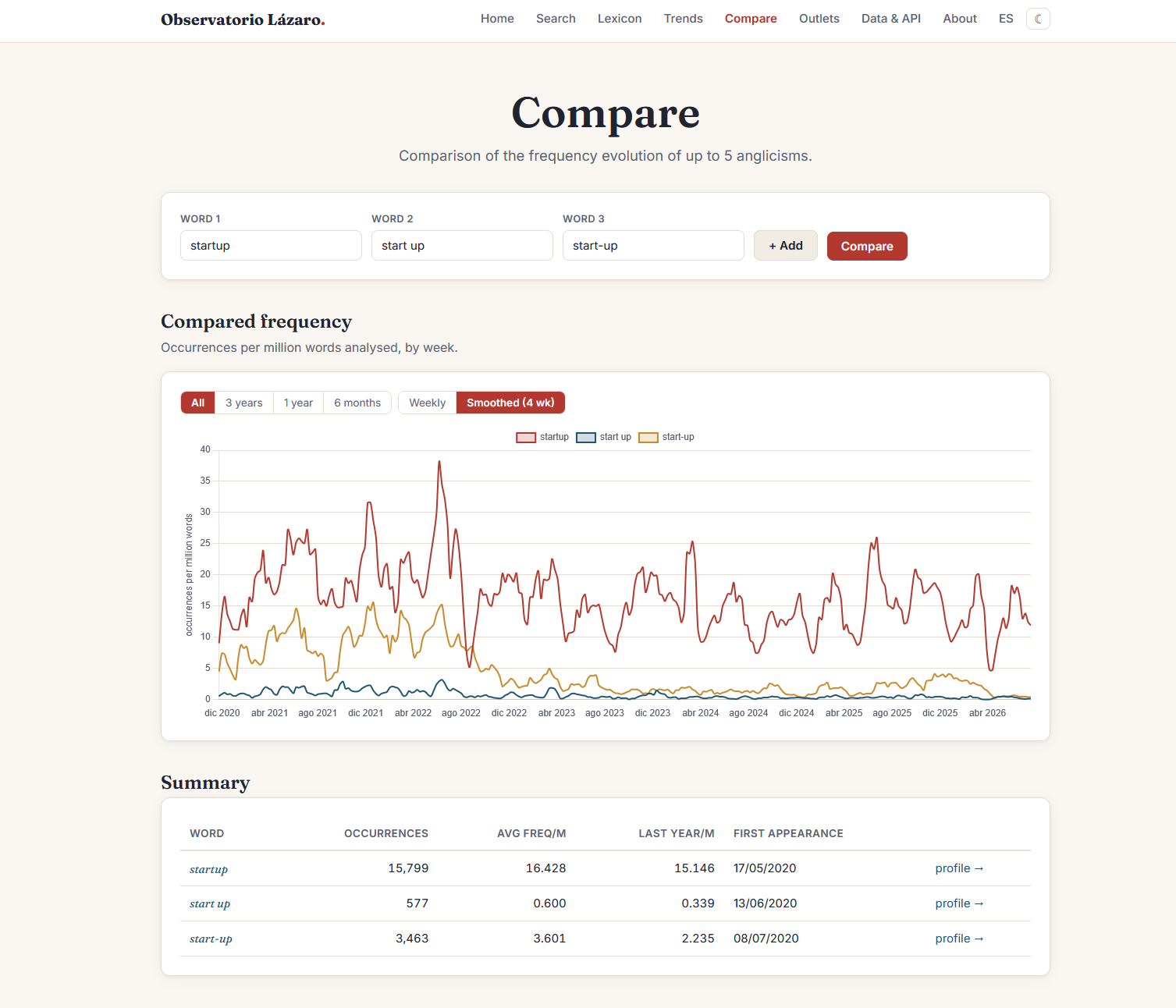}
    \caption{Comparison between competing anglicisms: \emph{startup}, \emph{start up}, \emph{start-up}.}
    \label{fig:startup}
\end{figure}

Five tools sit alongside the front page. The \emph{search} interface queries occurrences and their context sentences, with filters on outlet, section, date range and language label. The \emph{lexicon} browses the inventory of recorded borrowings. The \emph{trends} view ranks borrowings over a configurable window (the last week, month, quarter, year or three years) under three headings: the most frequent, those whose frequency in the period most exceeds their own historical average, and those detected for the first time. The \emph{comparison} tool plots the frequency series of up to five lemmas against one another, which is the form in which competition between rival borrowings is most legible (\ex{running} against \ex{footing}, or \ex{streaming} against \ex{podcast}) (see Figure~\ref{fig:startup}). The \emph{outlets} view ranks every monitored publication by borrowing density per million published words, reporting for each the number of articles, the running-token count and the raw number of detections, with a parallel ranking by section.

\begin{figure}[htbp]
\centering
\includegraphics[width=0.75\textwidth]{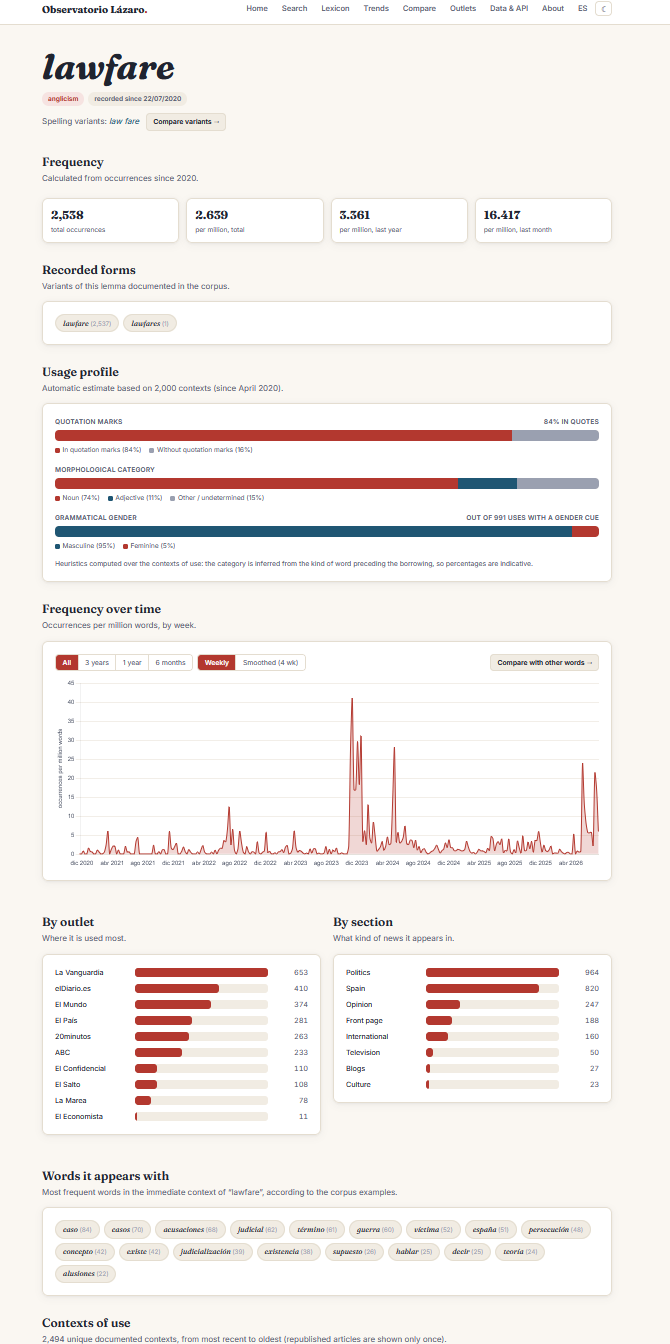}
\caption{Upper part of the lexical entry page for \ex{lawfare}, in the English interface.}\label{fig:entry}
\end{figure}

The lexical entry page is the part of the site closest to a dictionary entry, and it aggregates everything the observatory holds on one borrowing (Fig.~\ref{fig:entry}). For each lemma it reports the total number of occurrences and the normalized rate per million words over several periods; a frequency series, weekly or smoothed over four weeks; the surface forms grouped under the lemma with their individual counts; distributions by outlet and by section; the words that most often occur next to it; other borrowings related to it or containing it; and the full set of concordances, sortable by form, outlet, section and date. Listing the surface forms individually is what allows a user to inspect the composition of a lemma group rather than take the normalization on trust (Section~\ref{sec:pipeline-post}), and it is also what makes competing spellings visible (as with \ex{coming of age} beside \ex{coming-of-age}, or the English and Spanish plurals of \ex{celebrity}). The page additionally displays an automatically derived usage profile, giving the proportion of occurrences inside quotation marks and an assignment of part of speech and grammatical gender. 

Programmatic access is provided by a public JSON API, documented on the site's data page and summarized in Table~\ref{tab:api}. Responses are UTF-8 encoded and rate-limited to between twenty and sixty requests per minute depending on the endpoint, with excess requests returning HTTP~429. For bulk use the same page offers the database as monthly CSV files covering April 2020 to the present, so that a user can reconstruct the full record without paging through the API; a snapshot of the database is also available via Zenodo \citep{alvarez_mellado_2026_21721949}: searches run in the browser export up to 5{,}000 rows. Because the site is updated daily while the analyses reported in this paper are computed on a snapshot taken on 24 July 2026, figures retrieved from the site will exceed those reported here by whatever has accumulated since. Finally, the \texttt{pylazaro} library allows the detector itself to be run over new text, and all code is released as open source in the project repository.\footnote{\url{https://github.com/lirondos/lazaro}}

\begin{table}[htbp]
\centering
\caption{Public API endpoints. All return JSON; paths are relative to \texttt{https://observatoriolazaro.es/api/}.}\label{tab:api}
\begin{tabular}{@{}llp{5.2cm}@{}}
\toprule
Endpoint & Parameters & Returns \\
\midrule
\texttt{palabra.php}     & \texttt{word} & Full profile of one lemma: occurrences, normalized frequencies, surface forms, weekly series, distribution by outlet and section \\
\texttt{buscar.php}      & \texttt{q}; optionally \texttt{mode}, \texttt{media}, \texttt{section}, \texttt{from}, \texttt{to}, \texttt{lang}, \texttt{p} & Occurrences with their context sentences, paginated at 25 per page \\
\texttt{comparar.php}    & \texttt{words} (up to five) & Parallel frequency series for several lemmas \\
\texttt{tendencias.php}  & \texttt{tipo}, \texttt{periodo}, \texttt{dias} & Rankings by frequency, by rise over historical average, or by first detection \\
\texttt{sugerencias.php} & \texttt{q} & Lemma autocompletion \\
\botrule
\end{tabular}
\end{table}

\section{The resource: database description}\label{sec:resource}

\subsection{Data model}\label{sec:resource-model}

The database is organized around three related tables (Table~\ref{tab:schema}). The occurrences table stores one row per detected borrowing token; the \emph{index} table stores one row per borrowing type; and the articles table stores one row per processed article, providing the denominators required to normalize frequencies. A snapshot of the database has been publicly released in Zenodo \citep{alvarez_mellado_2026_21721949}.

\begin{table}[h]
\centering
\caption{Principal fields of the L\'azaro database (abbreviated).}\label{tab:schema}
\begin{tabular}{@{}lp{9.2cm}@{}}
\toprule
Table & Key fields \\
\midrule
Occurrences & borrowing (surface form), lemma, language, outlet, section, date, context (sentence), start/end token, in-headline flag, in-quotation flag \\
Index (per type) & borrowing, lemma, language, occurrences (total), first-attestation date, hapax flag \\
Articles & url, headline, date, outlet, section, running-token count \\
\botrule
\end{tabular}
\end{table}

\subsection{Size and coverage}\label{sec:resource-size}

Table~\ref{tab:coverage} summarizes the resource by year. It grows steeply: the annual volume of processed text rises from 40~million running tokens in 2020 to over 220~million in 2025, reflecting both continuous accumulation and the mid-2022 expansion of the outlet set. In total the resource records 2{,}007{,}647 borrowing tokens (68{,}424 distinct case-folded lemmas) in 1{,}880{,}377 articles over 993.4~million running tokens.\footnote{Throughout, `tokens' (or `running tokens') means \texttt{spaCy} tokens, the unit in which the text is processed and in which the corpus size is recorded. These include punctuation marks, which are tokenized separately: \ex{spin-off}, for instance, is three tokens (\ex{spin}, \ex{-}, \ex{off}). Densities expressed per token are therefore slightly lower than the equivalent rate per orthographic word, by the share of punctuation in the text (of the order of 10--15\% in news prose).} The overall density is approximately 2{,}020 borrowing tokens per million tokens, or (in more intuitive terms) about two unassimilated anglicisms per thousand tokens of Spanish press prose, roughly one every 500 tokens. Section~\ref{sec:eval-audit} shows that this estimate is stable under correction for detection error. 

\begin{table}[h]
\centering
\caption{Coverage of the resource by year (2020--2026; 2026 partial, to 24 July).}\label{tab:coverage}
\begin{tabular}{@{}lrrr@{}}
\toprule
Year & Articles & Running tokens & Borrowing tokens \\
\midrule
2020 & 131{,}391 & 39{,}529{,}071 & 79{,}010 \\
2021 & 286{,}289 & 87{,}294{,}941 & 199{,}442 \\
2022 & 303{,}836 & 121{,}524{,}903 & 271{,}036 \\
2023 & 293{,}020 & 186{,}089{,}991 & 384{,}343 \\
2024 & 348{,}365 & 217{,}885{,}418 & 443{,}659 \\
2025 & 352{,}904 & 227{,}817{,}827 & 433{,}797 \\
2026\,(part.) & 164{,}572 & 113{,}275{,}309 & 196{,}360 \\
\midrule
Total & 1{,}880{,}377 & 993{,}417{,}460 & 2{,}007{,}647 \\
\botrule
\end{tabular}
\end{table}

\subsection{Availability, licensing and reuse}\label{sec:resource-avail}
The live data are browsable and queryable through the website and its API\footnote{\url{https://observatoriolazaro.es/}}, and the complete 2020--2026 database can be downloaded in full. The detection model, the \texttt{pylazaro} library and the \textsc{coalas} corpus are available from their own repositories. The database contains the sentences in which borrowings were detected and the article headlines, but not full articles. The licenses applying to each component, and the status of this source-derived text, are set out in Appendix~\ref{app:data_availability}. A citable, versioned snapshot of the exact 2020--2026 database described here is available with a persistent identifier via Zenodo\footnote{\url{https://zenodo.org/records/21721949}} \citep{alvarez_mellado_2026_21721949}.

Following FAIR principles, the resource is findable (persistent website and repository), accessible (open web, API and library, plus a full download), interoperable (tabular records with documented fields) and reusable (open license and companion tooling).

\section{Resource evaluation}\label{sec:eval}

Because a monitoring resource is only as trustworthy as the detector that populates it, we evaluate it along three complementary axes: the detector's benchmark performance, the reliability of the annotation it learned from, and the precision of the data it actually produces in deployment.

\subsection{Model performance}\label{sec:eval-model}

On the \textsc{coalas} test set (58{,}997 tokens containing 1{,}239 English and 46 other-language borrowing spans), the deployed BiLSTM-CRF attains a span-level F1 of 0.86 for the borrowing class (precision 0.90, recall 0.82), the best result reported for the task \citep{alvarezmellado2022}. Table~\ref{tab:perf} breaks the result down by predicted language label. English borrowings, the overwhelming majority of the data, are detected well (F1~0.87), but the \textsc{other} class is detected poorly: precision is high (0.86) yet recall is only 0.13 (F1~0.23). The cause is a severe class imbalance in the training data, whose training split contains just 28 other-language borrowings against 1{,}493 English ones (the corpus-level counts of 123 and 3{,}038 given in Section~\ref{sec:related-detection} are distributed across the training, development and test splits), so the model rarely predicts \textsc{other} and tends to label any borrowing it detects as English. Two consequences follow for users: (i)~the resource is best understood as a monitor of \emph{anglicisms} specifically; and (ii)~the language tag stored with each occurrence is trustworthy for English but not for non-English borrowing, whose counts are substantially under-estimated (Section~\ref{sec:limits}).

\begin{table}[h]
\centering
\caption{Span-level detection performance on the \textsc{coalas} test set, overall and by language label \citep{alvarezmellado2022}.}\label{tab:perf}
\begin{tabular}{@{}lrrr@{}}
\toprule
Label & Precision & Recall & F1 \\
\midrule
All borrowings & 0.90 & 0.82 & 0.86 \\
English (\textsc{eng}) & 0.90 & 0.84 & 0.87 \\
Other (\textsc{other}) & 0.86 & 0.13 & 0.23 \\
\botrule
\end{tabular}
\end{table}

\subsection{Annotation reliability}\label{sec:eval-iaa}

The \textsc{coalas} corpus was manually annotated under a documented guideline. To assess its reliability, a sample of 9{,}110 tokens (450 sentences: 60\% from the test split, 20\% from training, 20\% from development) was doubly annotated by nine linguists; the mean inter-annotator pair-wise agreement, computed with Cohen's $\kappa$, was 0.91, well above the 0.8 threshold conventionally taken to indicate reliable annotation \citep{artstein2008,alvarezmellado2022}. This high agreement bounds the quality of the signal the detector learned from and provides a human ceiling against which its F1 (Section~\ref{sec:eval-model}) can be read.

\subsection{Precision of the deployed data}\label{sec:eval-audit}

The audit reported in this section was carried out on output of the current detector, and therefore characterizes the portion of the database produced from August 2022 onwards, which accounts for 79.5\% of all occurrences. The deployed precision of the superseded CRF model, and hence of the remaining 20.5\% of the database, has not been measured.

The F1 score of 0.86 reported in \citet{alvarezmellado2022} was measured on a curated test set in lab conditions. In order to assess the performance of the model when deployed on the observatory to identify borrowings on noisier real world data, we manually reviewed a subset of 1{,}000 spans from the observatory database (along with the context sentence they appeared in)\footnote{Due to copyright and storage limitations, the observatory only stores sentences where a borrowing was identified, which prevents us from evaluating recall retrospectively on the data already collected. The limitation is one of what has been stored rather than of the design: annotating a sample of complete articles as they are retrieved would measure recall directly without requiring the full text to be retained, and such a probe is planned (Section~\ref{sec:limits}).}. The spans were randomly selected but, given the Zipfian distribution of borrowings, we imposed the following two conditions when selecting the sample: 
\begin{enumerate}
    \item The 1{,}000 spans belonged to 1{,}000 distinct lemmas, so that high-frequency borrowings easily spotted by the model (such as \textit{look} or \textit{online}) did not dominate the sample.
    \item The 1{,}000 distinct spans were stratified along five frequency levels, to ensure that different types of borrowings were represented. Table \ref{tab:frequency} displays the number of spans per frequency tier.
\end{enumerate}

\begin{table}[h]
\centering
\caption{Number of spans selected for reannotation per frequency level. Total occurrences reports the total number of occurrences in the observatory}\label{tab:frequency}
\begin{tabular}{@{}llrl@{}}
\toprule
    Frequency tier & Total occurrences & No. of selected spans & Example  \\
\midrule
    Nonce & 1 & 250 & \textit{stress-free}, \textit{hyperlapse}\\
    Low & 2-10 & 250 & \textit{ziplock}, \textit{airball}\\
    Mid & 11-100 & 250 & \textit{journal}, \textit{developers} \\
    High & 101-1000 & 200 & \textit{coming-of-age}, \textit{carrot cake} \\
    Very high & $>$1000 & 50 & \textit{sold out}, \textit{mainstream}\\
\botrule
\end{tabular}
\end{table}

Out of the 1{,}000 manually annotated spans, 683 were true positives, which yields a raw precision of 0.68 on the reannotation sample. Because the sample is stratified by type across frequency tiers (Table~\ref{tab:frequency}), it deliberately over-represents the rare tail, so this raw figure is not itself a population estimate; we derive interpretable token- and type-weighted precisions from the per-tier results below (Section~\ref{sec:eval-audit}, Table~\ref{tab:audit}). Even so, it already points to a gap between processing clean, curated text under lab conditions and detecting borrowings on real-world data in the wild.

Table \ref{tab:errors} displays number of errors per type in the 1{,}000 spans. The most frequent type of error was caused by labeling as anglicism a span that was a non-English borrowing (such as \textit{bratwurst}, \textit{cocottes} or \textit{tofu}), French being the most frequent non-English donor language. This result is not surprising: results from \citet{alvarezmellado2022} already showed that the model was not reliable when identifying non-English borrowings. Boundary errors are the second most prevalent cause of error. Boundary errors account for partially retrieved multitoken spans (33; for instance, retrieving only \textit{dividend} in \textit{dividend recap}), spans that overlap with a true borrowing but also include non-borrowing tokens (9; \textit{brunch modernete}, when it should have retrieved only \textit{brunch}) and adjacent spans that were incorrectly fused into one single span (19; \textit{shopping bag beige}, instead of \textit{shopping bag} and \textit{beige}). 

In both cases (wrong language error and boundary error), the model correctly identified that there was a relevant span in the sentence, but it incorrectly labeled it as English origin or misidentified its boundaries. If we perform a more lenient evaluation in which a span is considered correct regardless of the assigned label and as long as there is some overlap between the predicted span and the goldstandard span (which is usually known as relaxed evaluation), precision goes up to 0.82, which is still below but closer to the precision scores reported on the official test set. Put differently, the two error types just described (wrong language and wrong boundaries) together account for 44\% of all errors: in almost half of the cases in which the model is wrong, a genuine unassimilated borrowing is nonetheless present in the sentence, and what fails is the language label or the delimitation of the span. This distinction matters for downstream use: a study of which anglicisms are used must apply the strict precision, whereas a study of how often borrowing occurs at all can rely on the relaxed precision.

The rest of the errors are due to foreign proper nouns, Spanish words (either odd-looking, already assimilated borrowings or words whose shape is equal to English words, such as \textit{horror}), English words not used as a true borrowing but as in metalinguistic discourse, codeswitches and other interferences (such as when quoting what someone said in English). Interestingly, 20 of the errors were caused by ill-tokenized URLs, hashtags, usernames and other Internet symbols, which should have been handled during preprocessing by simple heuristics. Other phenomena that confused the model and were a source of errors include typos, scientific names, acronyms and demonyms.
\begin{table}[h]
\centering
\caption{Number of true positives, false positives and error types on the selected 1{,}000 spans}\label{tab:errors}
\begin{tabular}{@{}lrl@{}}
\toprule
    Error type & Number & Example \\
\midrule
    True positives & 683 & \\
    WRONG LANGUAGE & 79 & \textit{boutique}, \textit{tofu}\\
    BOUNDARY ERROR & 61 & \textit{[pay to] play}\\
    PROPER NOUN & 50 & \textit{DSport}\\
    SPANISH WORDS & 40 & \textit{tiki-taka}, \textit{post-covid}, \textit{horror}\\
    METALINGUISTIC USAGE & 30 & \textit{yeah}\\
    URLs, EMAIL, HASHTAGS & 20 & \textit{change.org}, \textit{googlemail} \\
    CODESWITCHES & 15 & \textit{starring}\\
    OTHER & 22 & \textit{PVP}, \textit{vol.}, \textit{dB}, \textit{LOGSE}\\
\botrule
\end{tabular}
\end{table}

Because the sample was stratified, we can assess precision per frequency level. Table~\ref{tab:audit} shows that the model's worst result is obtained on nonce borrowings (those attested only once in the observatory), with a precision of 0.54: of all borrowings detected only once, roughly half are errors.

These errors are, however, of the same kinds as those observed in the sample as a whole. Of the 115 errors in the nonce tier, 47 (40\%) are language or boundary errors, that is, cases in which a borrowing is in fact present in the sentence but has been given the wrong language label or the wrong span. Only the remaining 60\% are cases in which no borrowing is present at all. Applied to a strict precision of 0.54, this means that about 25\% of one-off detections are not borrowings, while the other 75\%  do mark a real borrowing, even if it is mislabeled or misdelimited: the relaxed precision for the nonce tier is therefore approximately 0.72. The distinction matters for reuse, since a study of which anglicisms are used requires the strict figure, whereas a study of how often borrowing occurs at all can rely on the relaxed one.

Precision rises sharply with frequency, reaching 0.94 for the very-high tier: borrowings detected more than a thousand times are correct 94\% of the time. The real challenge, then, lies not in the frequent core but in the long tail of rare borrowings.

These per-tier figures also let us correct the raw sample precision of 0.68, which the stratified design inflates towards the tail. Weighting each tier's precision by the tier's actual share of the database (Table~\ref{tab:audit}) yields two interpretable population estimates. The token-weighted precision (the probability that a randomly retrieved occurrence is a genuine borrowing) is 0.87, close to the 0.90 obtained under lab conditions, because the token stream is dominated by high-frequency borrowings that the model detects reliably. The type-weighted precision (the probability that a randomly retrieved distinct borrowing is genuine) is much lower, 0.59, because the type inventory is dominated by the error-prone rare tail. This contrast is the crux of the resource's reliability profile: frequency- and trend-based analyses rest on data nearly as clean as the benchmark, whereas analyses that reach into the rare tail (e.g.\ neology or hapax studies) must contend with a substantially higher error rate and should apply the per-tier precisions reported here.

These figures also allow the resource's headline density estimate to be error-corrected. The raw detection rate is 2.02 borrowings per thousand tokens (Section~\ref{sec:resource-size}). Discounting false positives by the token-weighted precision (0.87) gives 1.76 genuine anglicisms per thousand tokens actually captured; further scaling by the detector's recall (0.82 on the test set) to account for borrowings that were missed gives an estimated true rate of 2.14 per thousand. The two error sources thus largely offset each other, and the estimate is stable at approximately two unassimilated anglicisms per thousand tokens of Spanish press prose (roughly one in every 500 tokens) across all three variants. This should be read as an order-of-magnitude estimate rather than a point value, since the recall correction assumes that test-set recall transfers to deployed data, which we cannot verify directly (Section~\ref{sec:limits}).

\begin{table}[h]
\centering
\caption{Precision audit per frequency tier, with each tier's share of the type inventory and of the token stream. Weighting the per-tier precisions by these shares yields a token-weighted precision of 0.87 (a randomly retrieved occurrence) and a type-weighted precision of 0.59 (a randomly retrieved distinct borrowing).}\label{tab:audit}
\begin{tabular}{@{}lrrr@{}}
\toprule
Frequency tier & Precision & \% of types & \% of tokens \\
\midrule
Nonce & 0.54 & 58.7 & 2.0 \\
Low & 0.63 & 31.1 & 3.9 \\
Mid & 0.75 & 7.9 & 8.1 \\
High & 0.77 & 1.9 & 19.2 \\
Very high & 0.94 & 0.4 & 66.8 \\
\botrule
\end{tabular}
\end{table}

\section{Statistical overview of the resource (2020--2026)}\label{sec:overview}

We close with a first characterisation of the resource, both to demonstrate its research value and to give prospective users a sense of its contents.

\subsection{Distribution of most frequent anglicisms}\label{sec:overview-scale}

\begin{figure}[h]
\centering
\includegraphics[width=0.72\textwidth]{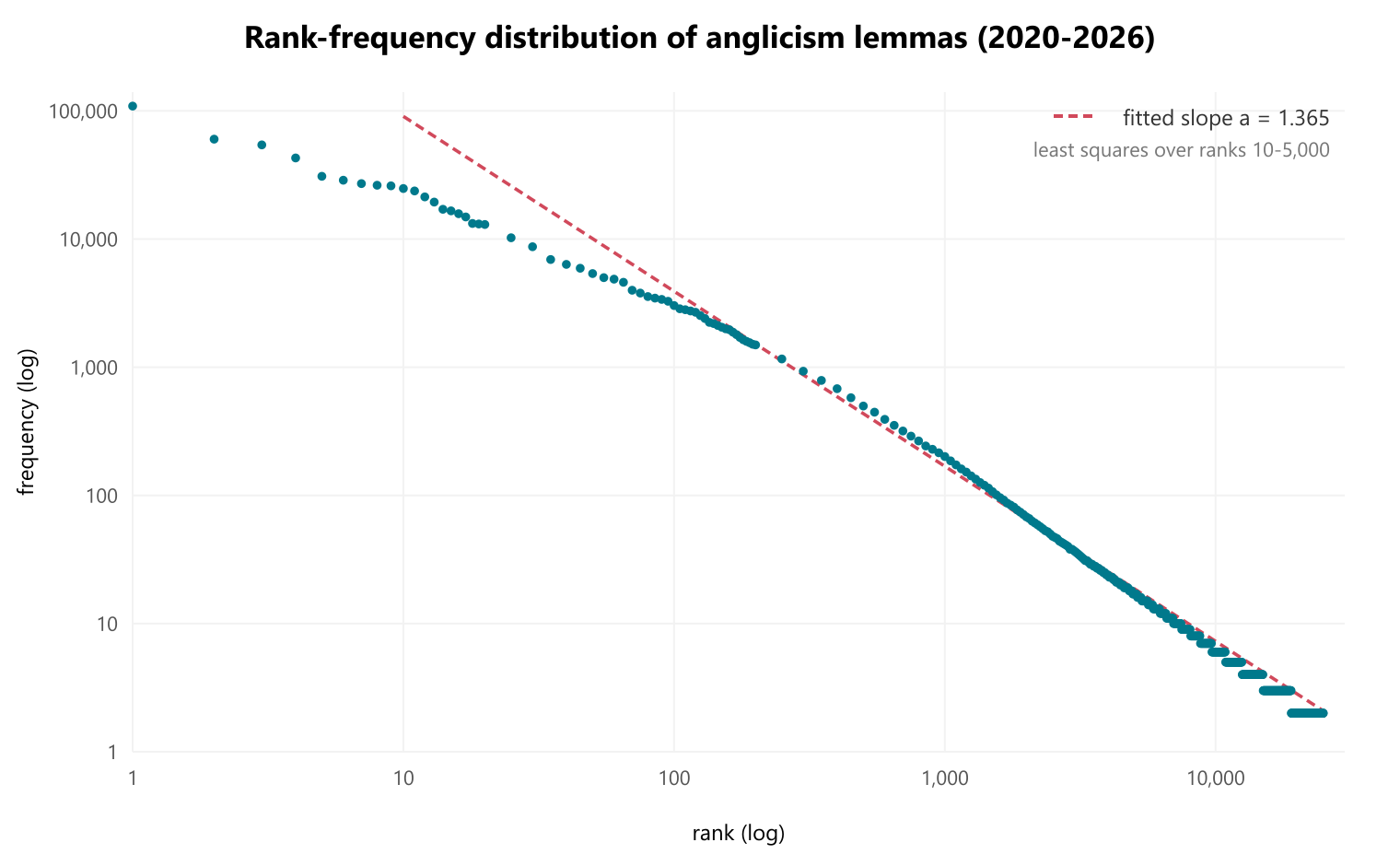}
\caption{Rank--frequency distribution of anglicism lemmas (log--log); dashed line, fitted Zipfian slope $-1.365$.}\label{fig:zipf}
\end{figure}

The anglicism vocabulary is dominated by a small high-frequency core over a very long tail. Counted over occurrences, the ten most frequent lemmas account for 21\% of all tokens and the top hundred for 51\%. Counted over the vocabulary, those hundred lemmas are 0.15\% of the 68{,}424 distinct types, while 58.7\% of the types occur exactly once and together contribute only 2\% of the tokens (53.6\% once each tier is discounted by its measured precision, Section~\ref{sec:eval-audit}). The gap between the two is the skew itself: a handful of borrowings carries half the running text, and the majority of the recorded vocabulary is barely attested at all. Table~\ref{tab:top} lists the twenty most frequent lemmas; they are overwhelmingly the vocabulary of digital media, technology and everyday evaluative register (\ex{look}, \ex{online}, \ex{influencer}, \ex{app}, \ex{streaming}). The rank--frequency distribution (Fig.~\ref{fig:zipf}) is steep and heavy-tailed, with a log--log slope of $a=1.365$ over ranks 10--5{,}000. The fitted range excludes the first nine ranks, where the head is flattened by a number of high-frequency borrowings of comparable rank, and stops at 5{,}000 to stay clear of the region in which counts fall to single figures and rank ties become pervasive. 
The distribution is thus reliably heavy-tailed, which is the usual finding for word frequencies.

\begin{table}[h]
\centering
\caption{The twenty most frequent anglicism lemmas (case-folded), with total occurrences.}\label{tab:top}
\begin{tabular}{@{}llr@{\quad}llr@{}}
\toprule
rank & lemma & freq. & rank & lemma & freq.\\
\midrule
1 & \ex{look} & 108{,}915 & 11 & \ex{software} & 23{,}720\\
2 & \ex{online} & 60{,}145 & 12 & \ex{marketing} & 21{,}320\\
3 & \ex{influencer} & 54{,}257 & 13 & \ex{newsletter} & 19{,}423\\
4 & \ex{app} & 42{,}910 & 14 & \ex{thriller} & 17{,}064\\
5 & \ex{reality} & 30{,}859 & 15 & \ex{shock} & 16{,}590\\
6 & \ex{streaming} & 28{,}765 & 16 & \ex{startup} & 15{,}759\\
7 & \ex{ranking} & 27{,}069 & 17 & \ex{smartphone} & 14{,}870\\
8 & \ex{top} & 26{,}284 & 18 & \ex{casting} & 13{,}229\\
9 & \ex{podcast} & 26{,}004 & 19 & \ex{post} & 13{,}103\\
10 & \ex{show} & 24{,}781 & 20 & \ex{prime time} & 12{,}986\\
\botrule
\end{tabular}
\end{table}

\subsection{Borrowings as a productive open class}\label{sec:overview-prod}

\begin{figure}[h]
\centering
\includegraphics[width=0.72\textwidth]{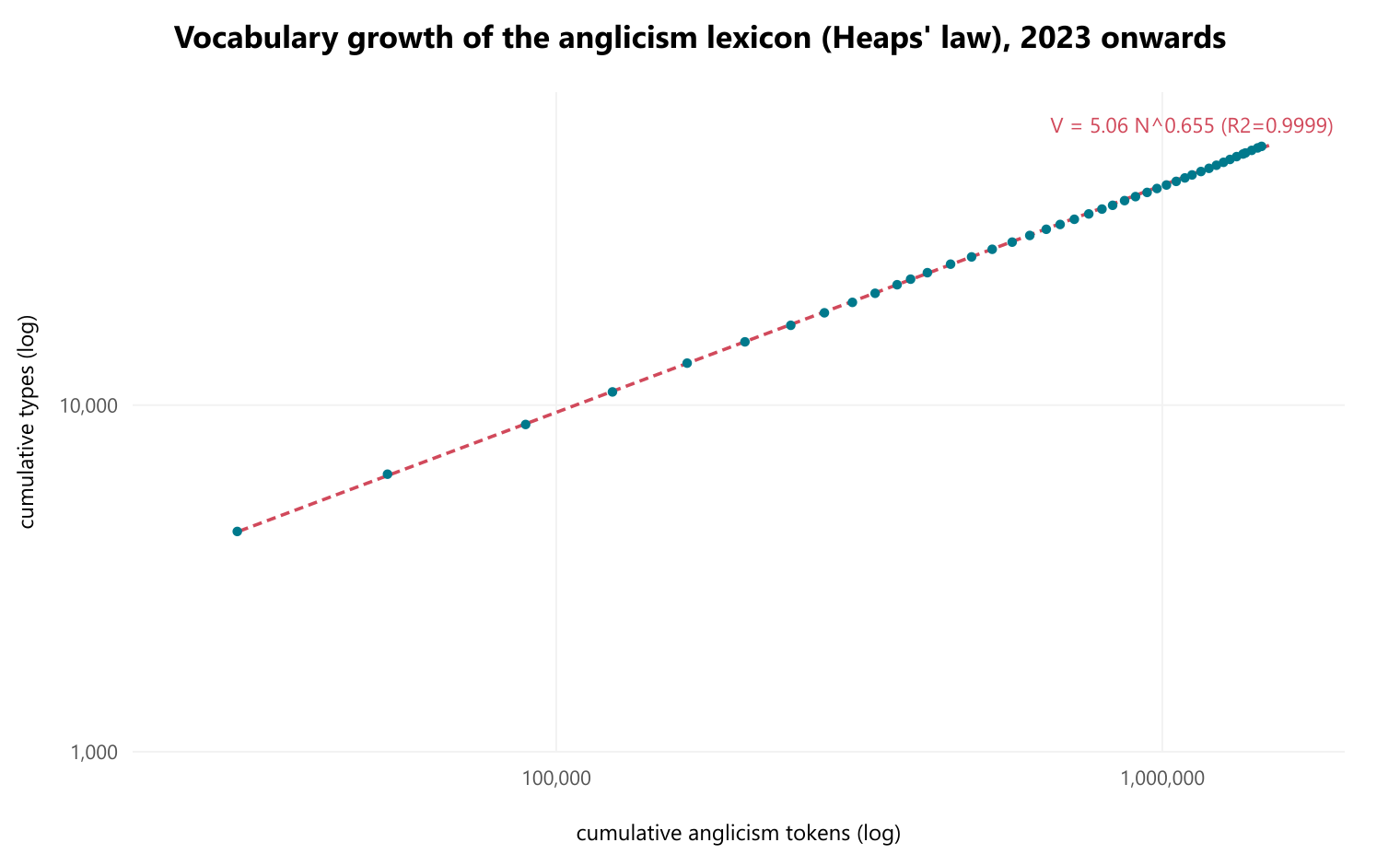}
\caption{Vocabulary growth (Heaps--Herdan law) from January 2023, the first full year after both the detector change and the outlet expansion: cumulative types against cumulative borrowing tokens (log--log), with fitted curve $V=5.06\,N^{0.655}$.}\label{fig:heaps}
\end{figure}

The vocabulary is not a closed inventory but a continuously renewed open class. Baayen's potential productivity $P=V(1)/N$ (the share of the token mass contributed by hapax legomena) is $0.020$ (roughly one new type per fifty borrowing tokens), Herdan's $C=\log V/\log N$ is $0.767$, and vocabulary growth follows the Heaps--Herdan law $V\propto N^{\beta}$ with $\beta=0.655$ ($R^2=0.999$; Fig.~\ref{fig:heaps}).

These are raw detection counts. Discounting every lemma by the measured precision of its frequency tier (Table~\ref{tab:audit}), and correcting the token stream on the same basis so that numerator and denominator are treated alike, gives $P=0.012$, $C=0.738$ and $\beta=0.611$ ($R^2=0.9999$). The correction is not a rescaling: it removes 46\% of the nonce types but only 6\% of the most frequent ones, so it changes the shape of the accumulation curve and not merely its intercept. All four constants fall, but the exponent remains well below one and the hapax share remains a majority of the inventory, so the vocabulary is appreciably less productive than the raw counts suggest while still behaving as an open and growing class.

The same curve can be read from the point of view of a reader rather than of the corpus, which gives a sense of how much text one has to go through before meeting a borrowing that is new to them. Over the period as a whole the observatory records 0.069 first attestations per thousand tokens, and the marginal rate at the current corpus size 
is 0.048. 
Discounted by the relaxed precision of the nonce tier (0.72, Section~\ref{sec:eval-audit}), this gives between 0.034 and 0.050 new borrowings per thousand tokens, that is, one previously unseen borrowing every 20{,}000 to 29{,}000 tokens. In other words, this amounts to one new borrowing every 38 to 56 articles of average length, or roughly one every four to six days for a reader who goes through ten articles a day; only about one in 51 borrowing encounters involves a word that is new to the reader. These figures should be read as a lower bound, since recall is unmeasured on the deployed data and novel forms are precisely where a static detector is most likely to fail (Section~\ref{sec:limits}).

\subsection{Borrowing across newspaper sections}\label{sec:overview-sem}

Because every occurrence records the newspaper \emph{section} in which it appears, the resource supports a data-driven view of where borrowing concentrates. Table~\ref{tab:section} reports, for the main sections, the borrowing density: occurrences per million running tokens, using each section's own token count as the denominator. 
The section labels are the ones the outlets themselves assign, as retrieved from the feed, and they are stored unmodified.

Borrowing is stratified by domain. Consumer, technology and entertainment sections lie above the corpus average of $\approx$2{,}020 per million: fashion shows the highest density at $\approx$10{,}500 per million, roughly five times the average, followed by technology and the women's section ($\approx$5{,}000), music, the men's section, television, cinema and lifestyle ($\approx$2{,}500--4{,}300). Hard-news and knowledge sections fall below the average, with science, international news, politics and national news between $\approx$590 and $\approx$880 per million. The spread across sections is therefore of more than an order of magnitude, from $\approx$10{,}500 down to $\approx$592 per million. The most frequent borrowings nonetheless include domain-general register words (\ex{look}, \ex{top}, \ex{boom}, \ex{shock}) that recur across all sections, so borrowing is partly stylistic and not only terminological.

\begin{table}[h]
\centering
\caption{Borrowing density by newspaper section, 2020--2026 (sections with $\geq$20{,}000 occurrences, sorted by density). Density is normalized by each section's own token count.}\label{tab:section}
\begin{tabular}{@{}lrr@{}}
\toprule
Section & Borrowings & Density (per M running tokens) \\
\midrule
Fashion          & 130{,}227 & 10{,}471 \\
Technology       & 126{,}073 &  5{,}041 \\
Women's          & 178{,}611 &  5{,}031 \\
Music            &  28{,}898 &  4{,}306 \\
Men's            &  27{,}460 &  3{,}350 \\
Television       & 168{,}755 &  2{,}937 \\
Cinema           &  22{,}299 &  2{,}841 \\
Lifestyle        & 154{,}263 &  2{,}541 \\
Celebrity        &  95{,}804 &  2{,}485 \\
Motoring         &  42{,}448 &  2{,}261 \\
Front page       & 277{,}014 &  2{,}061 \\
Economy          & 207{,}162 &  1{,}978 \\
Culture          & 138{,}316 &  1{,}830 \\
Sport            & 164{,}994 &  1{,}610 \\
Opinion          &  27{,}385 &  1{,}420 \\
Health           &  41{,}993 &  1{,}298 \\
Science          &  20{,}076 &    878 \\
International    &  51{,}219 &    700 \\
Politics         &  28{,}228 &    661 \\
National news    &  44{,}476 &    592 \\
\botrule
\end{tabular}
\end{table}

\subsection{Temporal dynamics}\label{sec:overview-time}



We now analyze borrowing density diachronically. We restrict our analysis to 2023 onwards, as it was the period of time when the same model was used and the collection of outlets remained stable\footnote{The detection model changed in 2022 from the CRF to the current BiLSTM-CRF model, and the collection of outlets monitored was expanded.}.


\begin{figure}[htbp]
\centering
\includegraphics[width=0.72\textwidth]{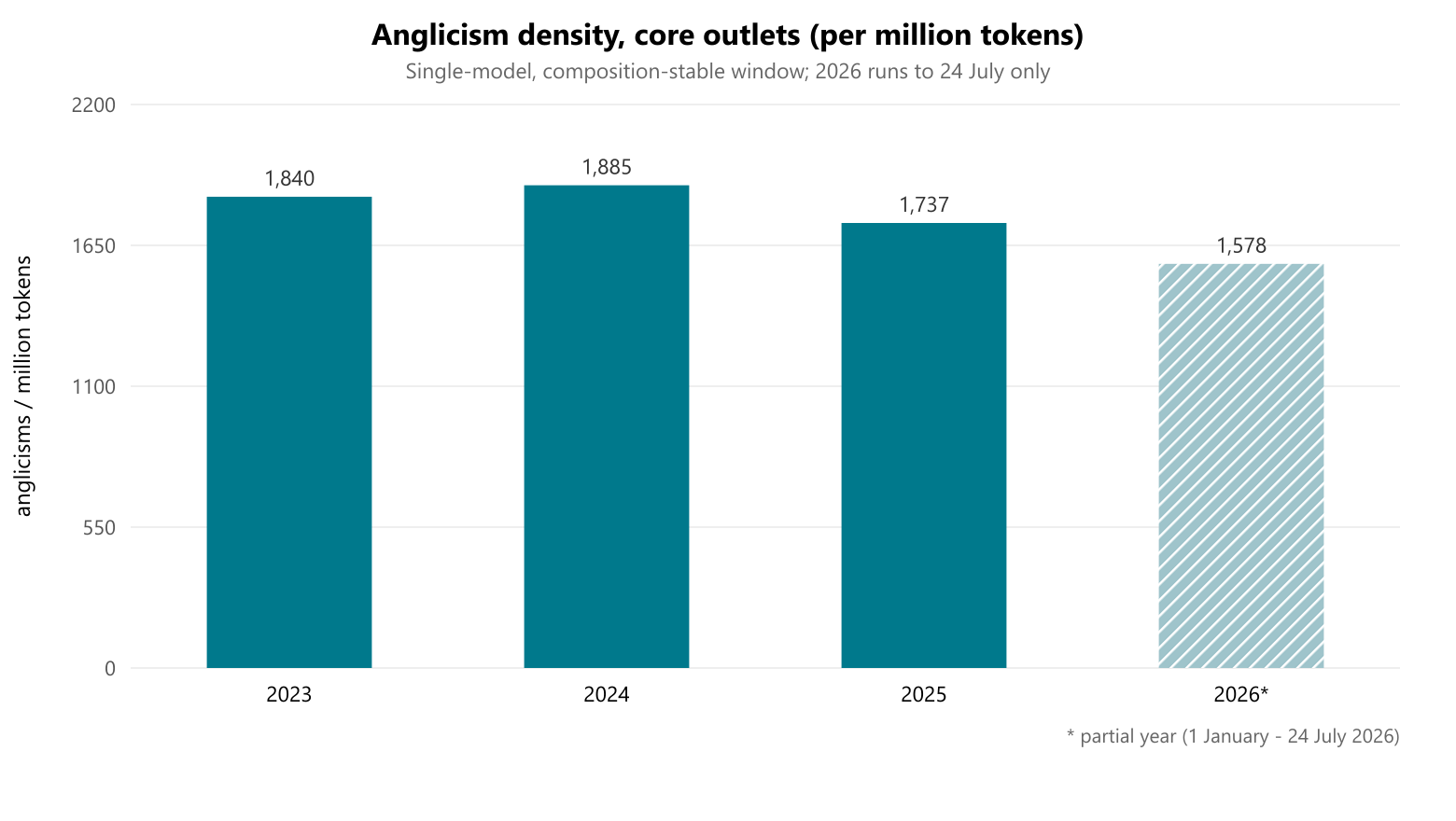}
\caption{Normalized anglicism density on the composition-stable core outlets, per million running tokens, by year. The series begins in 2023, the first full year after both the August 2022 detector change and the September 2022 outlet expansion; 2026 is hatched because it runs only to 24 July, and is excluded from the pooled rate quoted in the text.}\label{fig:density}
\end{figure}


Our data suggests that over the three years for which a single model has been in operation, the normalized rate of anglicism use in the Spanish press has remained stable (Fig.~\ref{fig:density}), with a rate of 1{,}817 borrowings per million tokens. This finding should however be taken cautiously: the detector is static, so its recall on borrowings that entered Spanish after training may decline over time. Because the observatory stores only sentences containing a detection, recall cannot be measured directly on the deployed data (Section~\ref{sec:limits}). The flatness reported above is therefore best read as an upper bound on any decline and a lower bound on any rise.

\section{Applications and reuse}\label{sec:apps}

Observatorio L\'azaro is designed for reuse across communities. For corpus and theoretical linguistics, its diachronic, frequency-annotated record supports the study of borrowing diffusion, productivity and establishment.
For computational linguistics, the deployed data and the \texttt{pylazaro} library provide training and evaluation material for borrowing and code-switching detection. For lexicography and language planning, the continuously updated lexicon is a candidate-detection feed for dictionaries of neologism. For teaching, the searchable interface offers attested, dated examples of contemporary borrowing. What these uses have in common is that they depend on observing the phenomenon as it occurs, at scale and with open access to the underlying data.

The content of the website and social media accounts provide a social dimension to the project geared towards a non-specialized public. In this regard, Observatorio L\'azaro seeks to contribute to the public conversation about language change and anglicism usage from a descriptivist perspective. We hope that the development of Observatorio L\'azaro can contribute to raise awareness about the possibilities that machine learning has to offer to the study of language contact and corpus linguistics in general, and the process of lexical borrowing in particular.

Finally, the purpose of Observatorio L\'azaro is to analyze the usage of borrowings in the Spanish press. This project does not seek to promote or stigmatize the usage of borrowings, or those who use them. The motivation behind our research is not to defend an alleged linguistic purity, but to study the phenomenon of lexical borrowing from a descriptive and data-driven point of view.

\subsection{Documented uptake}\label{sec:apps-uptake}

The resource is already in third-party use. To date, at least eleven independent publications have drawn on Observatorio L\'azaro's data, mostly within Hispanic linguistics and applied English studies, covering anglicisms in the language of economics \citep{de2023anglicismos}, Spanish toponymy \citep{lillo2022anglicismos}, the leisure and tourism domains \citep{10553_77321,10553_127747}, food and drink \citep{garcia2021anglicisms,lujan2022we,lujan2023anglicisms,lujan2023drink}, information technology \citep{10553_114908,NúñezNogueroles_Luján-García_2022}, and sports anglicisms and their metaphorical extensions in the digital press \citep{lujan2024political}. This uptake preceded any formal description of the resource, which is part of the motivation for documenting it here.

\section{Limitations}\label{sec:limits}

Five limitations should guide use. 
\paragraph{Detection is automatic and therefore imperfect.} The precision audit (Section~\ref{sec:eval-audit}) quantifies the error and shows it to be modest and concentrated in the low-frequency tail, but studies of rare types should apply the reported rates. Relatedly, the English/other language tag is reliable only for English (Section~\ref{sec:eval-model}): non-English borrowings are heavily under-detected and frequently mislabeled as English, so the resource should be used to study anglicisms rather than borrowing from other languages. 

\paragraph{The resource contains a mid-2022 discontinuity.} The detection model was upgraded (CRF~$\rightarrow$~BiLSTM-CRF) in August 2022 and the outlet set was expanded in September 2022, so measurements taken on either side of that boundary are not comparable. We therefore exclude cross-changeover comparisons altogether rather than qualify them, and restrict all diachronic statements to the single-model window 2023--2025 on the stable core outlets (Section~\ref{sec:overview-time}). 

\paragraph{Recall is unmeasured on the deployed data.} Because only sentences containing a detection are stored, borrowings the model misses leave no trace, so recall can be estimated only on the curated test set (0.82) and not in the wild. This matters especially for diachronic claims: the model has been unchanged since August 2022, and its recall on borrowings that entered Spanish after training may well be lower, so measured frequencies (and above all the apparent influx of new types) may increasingly under-represent genuine usage. Users making claims about emerging or rare borrowings over time should accordingly treat any measured trend as a lower bound on a rise and an upper bound on a decline (Section~\ref{sec:overview-time}). Periodic retraining on freshly annotated data, and a recall probe on full (undetected) sentences, are the natural remedies and are planned. 

\paragraph{Lemmatization has not been validated against human judgment.} Lemmatization affects around a tenth of lemma groups, which bounds its influence, but every type-level figure in Section~\ref{sec:overview} inherits this uncertainty. 

\paragraph{The resource is a monitor of the Spanish written press.} It reflects the register and topical priorities of news media in Spain and should not be treated as representative of the Spanish language as a whole.

\section{Conclusion}\label{sec:concl}

In this paper we have presented Observatorio L\'azaro, an observatory of anglicism usage in the Spanish press. The observatory automatically monitors the presence of English lexical borrowings in a collection of Spanish media sites in real time, and occupies a position between static borrowing dictionaries and one-off annotated corpora. The result is a database of over two million borrowings extracted from real context from Spanish press, the largest self-populating database of its kind. We are not aware of another continuously growing resource devoted specifically to monitoring borrowing usage in the wild. It is built on a previously published detector, and what the present paper contributes is the operational system, the diachronic database and the documentation of its evaluation and reuse.

The pipeline facilitates tracking anglicism frequency over time, and documents the incorporation of novel anglicisms along with their context, which can assist lexicographic work and corpus linguistics research. The observatory shows that computer-assisted methods for lexical borrowing detection can successfully be used to build real-world applications that can inform linguistic work in a systematic and data-driven fashion.


\begin{appendices}

\section{Outlet inventory}\label{app:outlets}

Table~\ref{tab:outlets} lists every outlet monitored by the observatory, with the date on which it entered the record and its contribution in running tokens and borrowing occurrences over the period 2020--2026. 


\begin{table}[h]
\centering
\caption{Outlets monitored by Observatorio L\'azaro, ordered by token contribution.}\label{tab:outlets}
\begin{tabular}{@{}llrrr@{}}
\toprule
Outlet & Type & First seen & Tokens (M) & Borrowings \\
\midrule
La Vanguardia & general & 2020-01-10 & 209.6 & 380{,}097 \\
20minutos & general & 2020-01-09 & 143.6 & 326{,}295 \\
El País & general & 2020-01-07 & 142.0 & 218{,}089 \\
El Confidencial & general & 2020-01-01 & 94.0 & 278{,}612 \\
El Mundo & general & 2020-01-22 & 85.0 & 132{,}098 \\
ABC & general & 2020-02-08 & 79.8 & 125{,}339 \\
ElDiario.es & general & 2020-01-01 & 77.1 & 126{,}710 \\
EFE & agency & 2020-06-18 & 16.5 & 12{,}144 \\
\midrule
Marca & sports & 2022-09-10 & 23.9 & 47{,}698 \\
El Economista & economy & 2022-09-12 & 21.5 & 42{,}214 \\
Elle & lifestyle & 2022-09-09 & 18.1 & 140{,}525 \\
Diez Minutos & celebrity & 2022-09-10 & 17.4 & 38{,}086 \\
Muy Interesante & science & 2022-09-17 & 11.9 & 13{,}474 \\
El Salto & politics & 2022-09-16 & 10.7 & 11{,}336 \\
Men's Health & lifestyle & 2022-09-10 & 8.2 & 27{,}460 \\
Fotogramas & cinema & 2022-09-10 & 7.8 & 22{,}299 \\
Jotdown & culture & 2022-09-10 & 6.7 & 14{,}099 \\
Rolling Stone & music & 2022-09-09 & 6.7 & 28{,}898 \\
Cinco Dias & economy & 2022-09-10 & 4.3 & 12{,}936 \\
La Marea & politics & 2022-09-11 & 3.2 & 3{,}615 \\
Agencia SINC & science & 2022-09-09 & 2.3 & 1{,}544 \\
Climatica & environment & 2022-09-11 & 2.1 & 2{,}644 \\
El Mundo Today & satire & 2022-09-12 & 0.6 & 1{,}174 \\
Público & general & 2022-09-17 & 0.1 & 99 \\
Expansion & economy & 2022-09-09 & 0.1 & 119 \\
Pikara & feminism & 2022-09-13 & 0.02 & 43 \\
\botrule
\end{tabular}
\end{table}

\section{Data and code availability}
\label{app:data_availability}
The data supporting the findings of this study are openly available in Zenodo as Observatorio L\'azaro snapshot v2.0, \url{https://doi.org/10.5281/zenodo.23948263} \citep{alvarez_mellado_2026_21721949}. The snapshot comprises four files: (i)~borrowing occurrences, with the surface form, the sentence containing it, source URL, outlet, section, publication date, predicted language label, token offsets, and a true/false value indicating whether the borrowing appeared in the headline or with quotation flags; (ii)~article records, with URL, headline, outlet, section, date and running-token count; (iii)~a per-form index, with lemma, total occurrences, first-attestation date and hapax flag; and (iv)~weekly token totals. Full article texts, author names and images are not stored or shared; the source articles remain available from their publishers through the URLs provided, subject to their continued availability and terms of access. Continuously updated data are available through \url{https://observatoriolazaro.es} and as monthly CSV files in the project repository \url{https://github.com/lirondos/lazaro}.

The CC BY 4.0 license of the Zenodo deposit, and the CC BY-SA 3.0 US license of the monthly files, apply only to the annotations, metadata and other material created by the author. They do not cover the sentences and headlines, which remain subject to the rights of their rightholders and over which no rights are granted. These extracts are included as citations for non-commercial scientific research under Art.~32 of the Spanish Intellectual Property Act: each is a brief extract from a published article, included to document a specific borrowing, and is identified by outlet and date and linked to the original article. Users are responsible for having an appropriate legal basis for any other reuse of these extracts.

RSS feeds are used only to detect new articles, whose text is then retrieved from the article page; neither the feeds nor outlet terms are relied on as permission, and outlet terms vary. Requests to remove a sentence or outlet can be sent to the corresponding author.

The \textsc{coalas} training corpus \citep{alvarezmellado2022}, which contains only text published under a Creative Commons license or used with permission, is available under CC BY-NC-SA 4.0 at \url{https://github.com/lirondos/coalas}. The detection model is available under CC BY 4.0 at \url{https://doi.org/10.57967/hf/11427} \citep{Alvarez-Mellado2026-qf}.
The \texttt{pylazaro} library is available under the MIT license at \url{https://github.com/lirondos/pylazaro}. The pipeline code is available under CC BY-SA 3.0 US at \url{https://github.com/lirondos/lazaro}.

\begin{table}[h]
\centering
\caption{license applying to each component of Observatorio L\'azaro. licenses cover the material created by the author; they do not extend to third-party sentences or headlines.}\label{tab:licenses}
\begin{tabular}{@{}p{3.0cm}p{4.2cm}p{4.6cm}@{}}
\toprule
Component & Location & license \\
\midrule
Zenodo snapshot, v1.0 (full downloadable database) & \url{https://doi.org/10.5281/zenodo.23948263} & CC BY 4.0 for annotations and metadata (source sentences and headlines remain subject
to the rights of their rightholders) \\
Monthly CSV files (downloadable database) & \url{https://github.com/lirondos/lazaro} & CC BY-SA 3.0 US for annotations and metadata (source sentences and headlines remain subject
to the rights of their rightholders) \\
\textsc{coalas} training corpus & \url{https://github.com/lirondos/coalas} & CC BY-NC-SA 4.0; text published under a Creative Commons license or used with permission \\
Detection model & \url{https://doi.org/10.57967/hf/11427} & CC BY 4.0 \\
\texttt{pylazaro} library & \url{https://github.com/lirondos/pylazaro} & MIT \\
Pipeline code & \url{https://github.com/lirondos/lazaro} & CC BY-SA 3.0 US \\
\botrule
\end{tabular}
\end{table}

\end{appendices}

\backmatter


\section*{Declarations}

\textbf{Competing interests.} The author declares no competing interests.\\
\textbf{Data availability.} See Appendix~\ref{app:data_availability}

\bibliography{sn-bibliography,additional,observatorio,biblio}

\end{document}